\documentclass[conference]{IEEEtran}
\IEEEoverridecommandlockouts
\usepackage{cite}
\usepackage{amsmath,amssymb,amsfonts}
\usepackage{algorithmic}
\usepackage{graphicx}
\usepackage{textcomp}
\usepackage{xcolor}
\usepackage{newfloat}
\usepackage{listings}
\newcommand{\lstbold}[1]{\textbf{#1}}

\usepackage{tabularx}
\usepackage{hyperref}
\usepackage{booktabs}
\usepackage{array}

\def\BibTeX{{\rm B\kern-.05em{\sc i\kern-.025em b}\kern-.08em
    T\kern-.1667em\lower.7ex\hbox{E}\kern-.125emX}}
    
\begin{document}

\title{GPT in Data Science: \\A Practical Exploration of Model Selection } 

\author{
	\IEEEauthorblockN{Nathalia Nascimento, Cristina Tavares, Paulo Alencar, Donald Cowan}
	\IEEEauthorblockA{\textit{David R. Cheriton School of Computer Science} \\
		\textit{University of Waterloo (UW)}\\
		Waterloo, Canada \\
		\{nmoraesd, cristina.tavares, palencar, dcowan\} @uwaterloo.ca}
}

\maketitle

\begin{abstract}

There is an increasing interest in leveraging Large Language Models (LLMs) for managing structured data and enhancing data science processes. Despite the potential benefits, this integration poses significant questions regarding their reliability and decision-making methodologies. It highlights the importance of various factors in the model selection process, including the nature of the data, problem type, performance metrics, computational resources, interpretability vs accuracy, assumptions about data, and ethical considerations. Our objective is to elucidate and express the factors and assumptions guiding GPT-4's model selection recommendations. We employ a variability model to depict these factors and use toy datasets to evaluate both the model and the implementation of the identified heuristics. By contrasting these outcomes with heuristics from other platforms, our aim is to determine the effectiveness and distinctiveness of GPT-4's methodology. This research is committed to advancing our comprehension of AI decision-making processes, especially in the realm of model selection within data science. Our efforts are directed towards creating AI systems that are more transparent and comprehensible, contributing to a more responsible and efficient practice in data science.


\end{abstract}


%

\section{Introduction}

At the 49th International Conference on Very Large Data Bases (VLDB), a panel led by Halevy et al. \cite{halevy2023will} posed an important question about the future of Data Science in the context of Large Language Models (LLMs). The growing role of LLMs in tasks such as database querying, query generation, and making inferences, as evidenced in recent studies \cite{chopra2023conversational,vert2023will,john2023datachat,troy2023enabling}, highlights this evolving landscape. 
The LangChain library \cite{LangChain} exemplifies this integration, enabling LLMs to work with various computational resources including data connectors, and has attracted significant attention, being utilized by over 30,000 developers in creating context-aware and reasoning applications.

\begin{figure*}[htb!]
\centering
\includegraphics[scale=0.50]{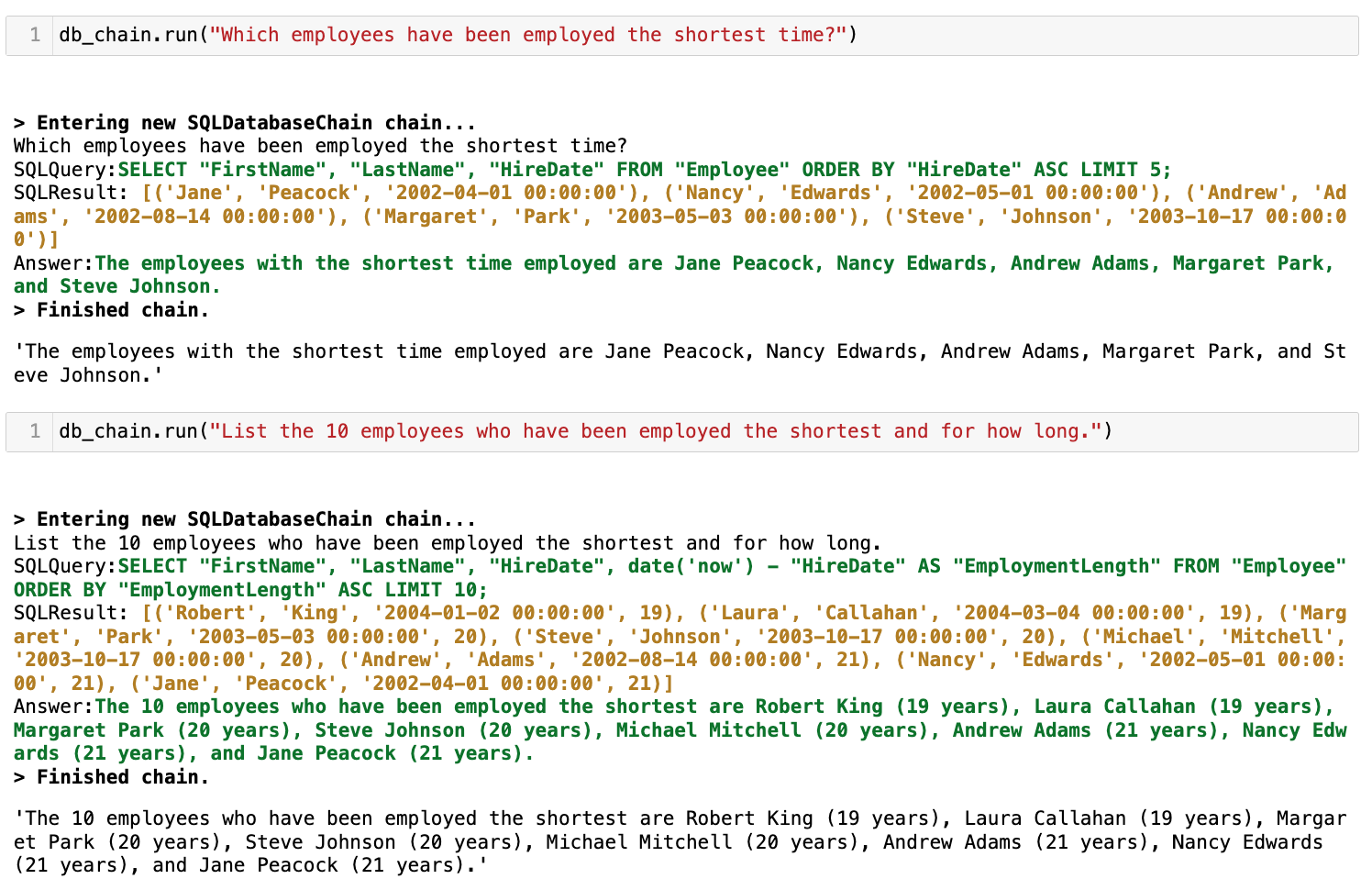}
\caption{LangChain with GPT-4: Failure in a simple SQL consultancy.}
\label{fig:failure}
\end{figure*}

Despite the increasing use of GPT models in various tasks \cite{nathalia2023artificial}, concerns about their reliability and decision-making processes remain, as illustrated in Figure \ref{fig:failure}. This figure demonstrates an inconsistency in GPT-4's query results when integrated with the LangChain framework, raising questions about the sources of such discrepancies (e.g. issues with LangChain, GPT-4, or the prompt structure).

\begin{figure*}[htb!]
\centering
\includegraphics[scale=0.50]{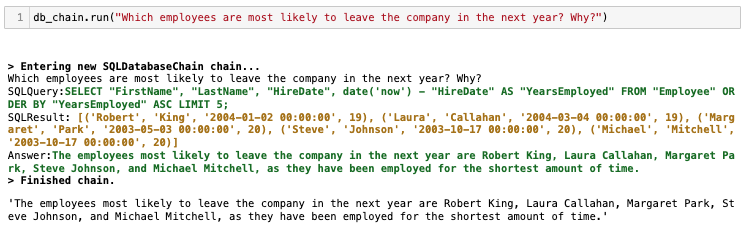}
\caption{LangChain with GPT-4: Exploring 'analytical' capabilities.}
\label{fig:analytical}
\end{figure*}

Further, examining GPT-4's analytical capabilities, as shown in Figure \ref{fig:analytical}, reveals challenges in understanding the underlying heuristics of its decisions. For instance, when querying about employees most likely to leave a company, GPT-4 focused on employment duration. This raises questions about the factors considered in its analysis and the potential need for more comprehensive data for improved analysis.

For GPT to be effectively utilized in decision-making within data science, it is essential to assess its explanatory capabilities. This paper concentrates on a critical aspect of data science decision-making: model selection. As indicated by Tavares et al. \cite{tavares2022adaptive}, the selection of machine learning models is influenced by various factors, including data attributes, prediction algorithm types, and requirements such as performance and bias. However, these influencing factors are often not explicitly represented, especially concerning bias detection. This paper delves into the nuances of model selection in Data Science, specifically focusing on the heuristics used by GPT-4. We aim to decipher these heuristics using the framework by Tavares et al. \cite{tavares2022adaptive}, which emphasizes a clear understanding and representation of the factors affecting model selection. Our approach seeks to enhance the transparency and efficiency of decision-making, contributing to a more explainable and effective data science practice.


We compare the heuristics used by GPT-4 with those proposed by Scikit-Learn engineers, as outlined by Tavares et al. \cite{tavares2022adaptive}. Using toy datasets, we examine GPT-4's decision-making process in model selection and its alignment with established heuristics. The paper addresses two key research questions: (RQ1) How can we capture and represent GPT's algorithm selection factors? (RQ2) How do GPT's heuristic outcomes compare to those of Scikit-Learn engineers? 

The paper is organized to offer a thorough understanding of GPT's role in Data Science. Section 2 provides a background on LLMs and their applications. Section 3 details our methodology for capturing and comparing heuristics. Section 4 presents our experimental findings, and Section 5 concludes the paper with a summary and future research directions.

\section{Background} \label{sec:background}

\subsection{LLM and GPT}

Large Language Models (LLMs) and Generative Pre-trained Transformers (GPT) are integral parts of AI's Natural Language Processing (NLP) realm. While LLM is a broad category encompassing models that predict word sequences and can be used for various tasks such as text generation and translation, GPT, developed by OpenAI \cite{OpenAI2023}, is a specific LLM type. GPT, renowned for generating text akin to human writing, undergoes extensive pre-training before fine-tuning for specialized tasks. In essence, GPT is a subclass of LLMs, but not all LLMs are GPT models. Other prominent LLM examples include BERT, RoBERTa, and XLNet.

 \subsection{Variability-aware model}

Variability modeling, widely explored for developing automated systems, refers to the process of identifying and representing the variabilities—or the ability of a product, item, or feature to change, evolve, or be customized—in order to harness automation opportunities. This concept is crucial in fields like software engineering, databases, and data warehouses, where understanding the common and unique aspects of a system is essential. Feature modeling, a key aspect of variability modeling, involves capturing the common and variable attributes of a system using abstract entities known as features. 

In this context, feature models stand out as intuitive and effective tools for representing the features of a variant-rich software system. They not only facilitate an overall understanding of the system but also support various stages of development, such as scoping, planning, development, variant derivation, configuration, and maintenance, thus contributing to the system's long-term success\cite{nevsic2019principles}. Feature models formally represent features, their relationships, and constraints. This approach enhances understandability, traceability, explainability, and maintenance of the system.

In our proposed reverse engineering approach, we use an automated method based on massive neural networks for decision-making. By extracting variability models from these automated systems, we aim to leverage the benefits of variability models—such as traceability, understandability, explainability, and maintenance—in a novel context. 

Furthermore, feature diagrams provide a structured, visual method for addressing the complexities in machine learning applications. These diagrams not only highlight the intricate relationship between modeling assumptions and algorithm choices but also underscore their impact on performance and other critical evaluation criteria, including fairness.

\subsection{LLM applied to Data Science}
Large Language Models (LLMs) are increasingly integral to data science, showcasing their potential in diverse areas such as data preprocessing, analytics, and even drug discovery \cite{vert2023will}. The study by Chopra et al. \cite{chopra2023conversational} highlights the use of LLMs in data science, particularly for tasks like data preprocessing and analytics. However, challenges arise in the interaction between data scientists and LLM-powered chatbots, especially in areas like contextual data retrieval, prompt formulation for complex tasks, and adapting generated code. These insights lead to the proposal of design improvements, including data brushing and inquisitive feedback loops, to enhance AI-assisted data science tools.

Tools like DataChat AI \cite{john2023datachat}, Anaconda Assistant, and Databricks Assistant, LLMs enable data scientists to engage in conversational interactions about their data. This interaction includes asking follow-up questions and receiving context-specific responses, greatly enhancing the user experience and efficiency in data management and analysis.
Finally, Troy et al. \cite{troy2023enabling} focus on enabling generative AI to produce SQL statements. They propose a tool that generates syntactically valid language sentences and integrate AI algorithms for semantic guidance, demonstrating the capability of LLMs in generating structured queries for specific purposes, like detecting cyber-attacks.

These studies collectively underscore the expanding role of LLMs in data science, from enhancing analytics and data management to contributing significantly to fields like drug discovery. However, the challenges, including effective communication with AI assistants and adapting AI-generated solutions to specific contexts, remain crucial areas for further development.

\section{Approach: Capturing and Representing Heuristics Behind GPT's Decision-Making Process} \label{sec:approach}
In this section, we explore the process of reverse engineering the decision-making heuristics of GPT (Generative Pre-trained Transformer) using both bottom-up and top-down approaches. Our goal is to deduce the underlying principles and factors that GPT considers when making decisions about data science model selection, given a specific dataset. This exploration is critical for understanding the capabilities and limitations of GPT in the context of data science and machine learning. The following subsections detail the methodologies and findings of our investigation.

\subsection{Top-down Approach}
In the top-down approach, we directly posed a question to GPT to elicit its criteria for selecting a data science model based on given dataset characteristics. The question was formulated as: “Given a dataset, which modeling assumptions/factors do you consider to selecting a data science model?" This query aimed to uncover the explicit decision-making criteria used by GPT. However, it's important to note that the response presented below may not necessarily reflect the actual decision-making process used by GPT internally; it could represent an idealized or generalized answer suitable for the given question.


\begin{lstlisting}[ frame=single, breaklines=true]
1. Nature of the Data:
Size: Larger datasets might require more robust models.
Type: Categorical, numerical, time-series, text, or image data can dictate the type of model.
Quality: The presence of missing values, outliers, or noise in the data can influence the choice of model and preprocessing steps.

2. Problem Type:
Classification vs Regression: For predicting categorical outcomes, classification models are used, whereas regression models are chosen for continuous outcomes.
Clustering: If the task involves finding groups within the data, clustering algorithms like K-means or hierarchical clustering are appropriate.
Dimensionality Reduction: Techniques like PCA are used when reducing the number of features is necessary.

3. Performance Metrics:
Different models are evaluated based on metrics like accuracy, precision, recall, F1 score for classification tasks, and mean squared error, mean absolute error for regression tasks.

4. Computational Resources:
The availability of computational resources can dictate the choice of model. 

5. Interpretability vs Accuracy:
Some models like decision trees or linear regression are more interpretable but might offer less accuracy compared to complex models like neural networks.

6. Assumptions About Data:
Certain models come with inherent assumptions. For example, linear regression assumes a linear relationship between features and the target variable.

7. Overfitting and Underfitting Concerns:
Models should be complex enough to capture the underlying patterns in the data but not so complex that they overfit.

8. Training Time:
The time it takes to train a model can be a critical factor, especially in environments where models need to be updated frequently.

9. Domain Knowledge:
Understanding the domain can help in choosing features and models that are more likely to succeed. For instance, in healthcare, certain models might be preferred due to their ability to handle specific types of medical data.

10. Previous Experiences and Benchmarks:
Often, previous experiences with similar datasets or benchmarks in literature can guide the choice of the model.

11. Ethical Considerations:
Models should be chosen and designed keeping in mind ethical considerations like fairness, privacy, and transparency, especially in sensitive fields like healthcare and agriculture.
\end{lstlisting}

\subsection{Bottom-up Approach}
To complement the top-down analysis, we employed a bottom-up approach, which involved practical applications of GPT’s decision-making process. We selected three different toy datasets, each with a clearly defined target variable. For each dataset, we asked GPT to recommend the most suitable machine learning model(s) to address the specific problem presented by the dataset. The prompt was structured as follows: ``Given the dataset \{x\}, with the target variable in column \{column y\}, and with the objective of \{problem description z\}, identify the most suitable machine learning model(s) to solve this issue. Explain your choice(s) and the underlying modeling assumptions and factors that guided your decision. Outline your decision-making process in detail. If multiple models are viable, rank them in order of preference, and describe the criteria for transitioning from one model to another in the evaluation process." 


\subsubsection{Dataset: Heart Failure Prediction}
This application aims to predict patient survival in heart failure scenarios using clinical data \cite{chicco2020machine}. The dataset comprises 299 patient records, each featuring 12 clinical attributes, such as age, sex, diabetes presence, and serum creatinine levels. Additionally, it includes a binary target variable, `death\_event', indicating patient mortality within an average follow-up period of 130 days \cite{chicco2020machine}.

Prompt: ``Given the attached dataset \{heart\_failure\_clinical\_records\_dataset.csv\}, with the target variable in column \{`death\_event'\}, and the objective of predicting survival of patients with heart failure based on clinical information, identify the most suitable machine learning model(s) to solve this issue. Explain your choice(s) and the underlying modeling assumptions and factors that guided your decision. Outline your decision-making process in detail. If multiple models are viable, rank them in order of preference, and describe the criteria for transitioning from one model to another in the evaluation process."

As a result, GPT provided the following result:

\begin{lstlisting}[frame=single, breaklines=true, escapeinside={(*@}{@*)}]
(*@\lstbold{Understanding the Dataset and Target Variable:}@*)
The dataset contains 299 entries and 13 columns.
The target variable is DEATH_EVENT, a binary variable (0 or 1), indicating whether the patient survived or not.
Features include a mix of numerical and binary categorical data (e.g., anaemia, diabetes, high_blood_pressure, sex, smoking) and continuous variables (e.g., age, creatinine_phosphokinase, ejection_fraction, platelets, serum_creatinine, serum_sodium, time).
(*@\lstbold{Model Selection Criteria:}@*)
Handling Binary Classification: Given the binary nature of the target variable, models suited for binary classification are appropriate.
Interpretability vs. Predictive Power: Models that offer a balance between interpretability and predictive accuracy are preferred. This is crucial in healthcare applications.
Handling Mixed Data Types: The model should handle both continuous and categorical data effectively.
Robustness to Overfitting: Given the relatively small dataset size, models that are less prone to overfitting are desirable.
Computational Efficiency: Considering the dataset size, models that are computationally efficient are preferred.
(*@\lstbold{Proposed Models and Their Suitability:}@*)
Logistic Regression: A good baseline model for binary classification, offering interpretability and simplicity. It assumes linear relationships between the log-odds of the outcome and predictors.
Random Forest: An ensemble method that is good at handling a mix of data types and is less prone to overfitting. It provides importance scores for features but is less interpretable than logistic regression.
Gradient Boosting Machines (GBM): Similar to Random Forests, GBMs are powerful for classification tasks, handling mixed data types well, and often provide better predictive accuracy. However, they are less interpretable and more prone to overfitting.
Support Vector Machine (SVM): Effective in high-dimensional spaces and with different kernel functions, SVMs can model non-linear relationships. They are less interpretable and require careful tuning of hyperparameters.
(*@\lstbold{Ranking and Evaluation Process:}@*)
Logistic Regression: Start with this for its simplicity and interpretability.
Random Forest: If Logistic Regression underperforms, move to Random Forest for its balance between accuracy and overfitting resistance.
Gradient Boosting Machines: If further improvement is needed and the complexity is manageable.
Support Vector Machine: Consider if non-linear relationships are strongly suspected in the data.
(*@\lstbold{Transition Criteria:}@*)
Performance Metrics: Transition based on metrics like accuracy, precision, recall, and AUC-ROC.
Overfitting Signs: If a model shows signs of overfitting (e.g., high variance in cross-validation scores), consider switching to a more robust model.
Computational Constraints: If computational resources are limited, prioritize simpler models.

In summary, start with Logistic Regression for its simplicity and interpretability, then move to more complex models like Random Forest or GBM if needed, based on performance metrics and overfitting considerations. SVM can be considered for potential non-linear relationships.
\end{lstlisting}

\begin{figure*}[!ht]
	\centering
	\includegraphics[scale=0.80]{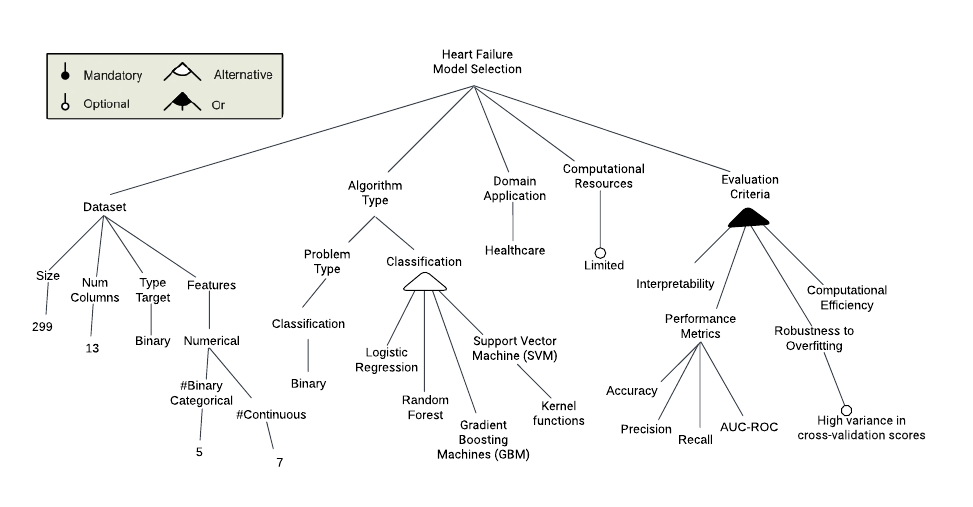}
	\centering
	\caption{Feature diagram representing GPT heuristics for model selection in the Heart Failure Dataset.}
	\label{fig:App01}
\end{figure*}

Based on this response, we could extract a list of factors that were considered for proposing a rank of the most-fit models. Following the GPT guidance, the feature model shown in Figure \ref{fig:App01} was developed.

Based on the GPT output and on the feature model we designed for this application, the key constraints influencing feature modeling include:
\begin{enumerate}
    \item Dataset Constraints:
    \begin{itemize}
        \item DatasetSize = 299 $\wedge$ NumberOfColumns = 13
        \item Target Variable Type = Binary (DEATH\_EVENT)
        \item FeatureTypes = (Numerical $\vee$ BinaryCategorical)
    \end{itemize}
    \item Algorithm Type Constraints:
    \begin{itemize}
        \item Problem Type = Classification
        \item Classification Type = Binary
        \item Algorithm Options = \{Logistic Regression, Random Forest, Gradient Boosting Machines, Support Vector Machine\}
    \end{itemize}
    \item Evaluation Criteria Constraints:
    \begin{itemize}
        \item EvaluationCriteria = (Interpretability $\vee$  PerformanceMetrics $\vee$  ComputationalEfficiency $\vee$  RobustnessToOverfitting)
        \item PerformanceEvaluation $\Rightarrow$ (Accuracy $\vee$ Precision $\vee$ Recall $\vee$ AUC-ROC)
    \end{itemize}
    \item Model-Specific Constraints:
    \begin{itemize}
        \item Logistic Regression $\Leftrightarrow$ (BinaryClassification $\wedge$ Interpretability $\wedge$ Simplicity)
        \item Random Forest $\Leftrightarrow$ (Robustness to Overfitting $\wedge$ Handling Mixed Data Types $\wedge$ FeatureImportanceScores $\wedge$ $\neg$Interpretability)
        \item Gradient Boosting Machines $\Leftrightarrow$ (High Predictive Accuracy $\wedge$ Handling Mixed Data Types $\wedge$ Prone to Overfitting)
        \item Support Vector Machine $\Leftrightarrow$ (Kernel Functions $\wedge$ High Dimensional Handling $\wedge$ Non-Linear Relationship Modeling $\wedge$ $\neg$Interpretability $\wedge$ HyperparameterTuning)
    \end{itemize}
    \item Computational Efficiency Consideration:
    \begin{itemize}
        \item Limited Computational Resources $\Rightarrow$ Prefer \{Logistic Regression, Other Simplified Models\}
    \end{itemize}
    \item Ranking and Transition Process Constraints:
    \begin{itemize}
        \item InitialModel $\Rightarrow$ LogisticRegression
        \item If Logistic Regression Underperforms $\Rightarrow$ Consider \{Random Forest, Gradient Boosting Machines\}
        \item If Robustness to Overfitting Not Sufficient and Computational Resources are Unlimited $\Rightarrow$ Consider Gradient Boosting Machines
        \item If Non-Linear Relationships Suspected and Computational Resources are Unlimited $\Rightarrow$ Consider Support Vector Machine
    \end{itemize}
\end{enumerate}

\subsubsection{Dataset: Diabetes Prediction}
The Diabetes Prediction dataset \cite{diabetes}, utilized for modeling the likelihood of diabetes in patients, integrates medical and demographic data alongside diabetes status. It encompasses variables such as age, gender, BMI, hypertension, heart disease, smoking history, HbA1c level, and blood glucose level. The dataset's primary aim is to facilitate the development of machine learning models for diabetes prediction, aiding healthcare professionals in identifying at-risk individuals and aiding research on diabetes-related factors.

Unique to this dataset are considerations for the complexity of relationships and the inclusion of Decision Trees and Neural Networks in the model selection process, factors not emphasized in the Heart Failure Prediction dataset. The dataset's categorical and numerical nature necessitates models adept at handling both data types, with the binary `diabetes' target variable framing it as a classification problem.

Key considerations include:

1. \textbf{Data Complexity}: Acknowledging potential non-linear relationships in the data, models capable of addressing complex data interactions are prioritized.

2. \textbf{Model Selection}: Decision Trees and Neural Networks are considered, in addition to Logistic Regression, Random Forests, and GBM. Decision Trees offer interpretability and manage non-linear data well, while Neural Networks excel in modeling intricate relationships, especially in larger datasets.

3. \textbf{Performance and Transition Criteria}: Model selection is guided by performance metrics (accuracy, AUC-ROC), with a focus on balancing interpretability and complexity. The transition between models is dictated by these metrics and signs of overfitting, ensuring model robustness and efficiency.

\begin{figure*}[!ht]
	\centering
	\includegraphics[scale=0.82]{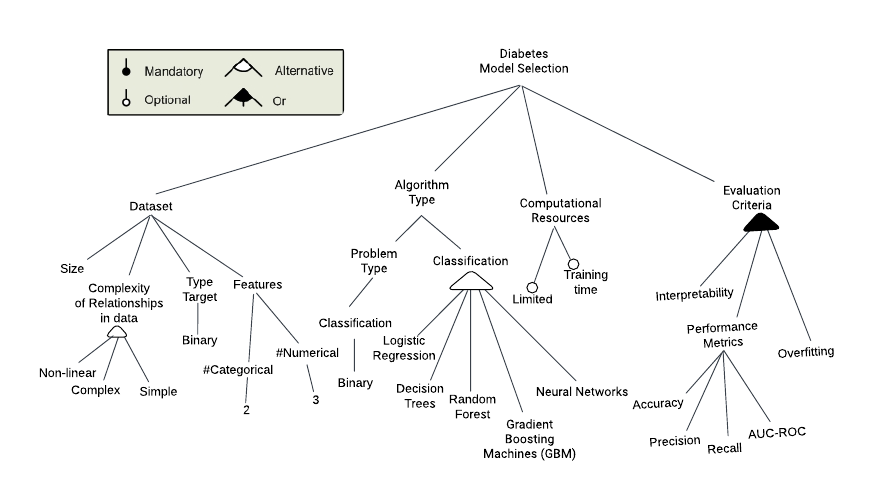}
	\centering
	\caption{Feature diagram representing GPT heuristics for model selection in the Diabetes Prediction Dataset.}
	\label{fig:App02}
\end{figure*}

The feature diagram (Figure \ref{fig:App02}) represents these heuristics, underscoring the unique aspects of the Diabetes Prediction dataset in model selection, unlike the Heart Failure dataset.

Ranking and Transition Criteria:
\begin{lstlisting}[frame=single, breaklines=true]
Logistic Regression: Start with this for a baseline. If the performance (e.g., accuracy, precision, recall) is unsatisfactory, move to more complex models.
Decision Trees: If logistic regression is insufficient and a more interpretable model is needed.
Random Forests: If the decision tree underperforms or overfits, this would be the next step for better accuracy and handling overfitting.
GBM: If further improvement is needed and interpretability is less of a concern.
Neural Networks: Consider this if the dataset is large and the above models do not provide satisfactory results, and if computational resources permit.

Transitioning from one model to another should be based on performance metrics relevant to the problem (like accuracy, AUC-ROC) and the complexity of the model, considering the trade-off between performance and interpretability. Regular cross-validation should be used to ensure the model is not overfitting.
\end{lstlisting}

\subsubsection{Dataset: Car's Price Prediction}
\begin{figure*}[!ht]
\centering
\includegraphics[scale=0.80]{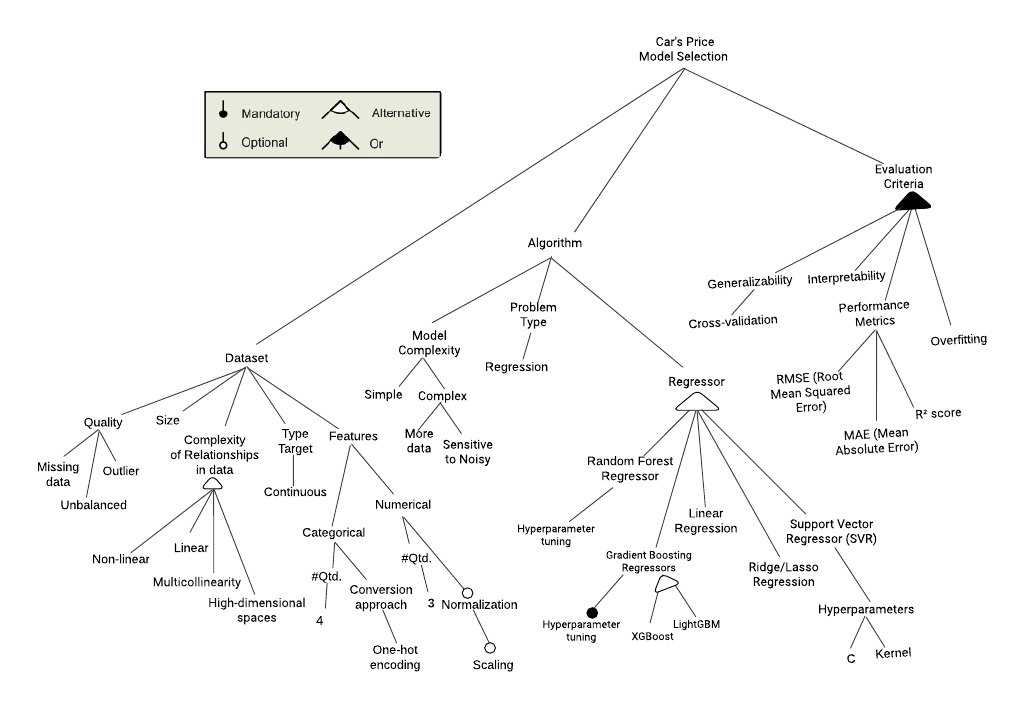}
\centering
\caption{Feature diagram to represent GPT heuristics for model selection in the Cars'Price Prediction Dataset.}
\label{fig:App03}
\end{figure*}

Based on the car price prediction dataset \cite{vehicle} and the nature of the regression problem, the selection of suitable machine learning models incorporates additional heuristics and factors, as depicted in Figure \ref{fig:App03}:
\begin{enumerate}
    \item Dataset Complexity (Simple vs. Complex Regressor): The complexity of relationships in the data is considered. Simple regressors are suitable for less complex relationships, while complex models are favored for more intricate data patterns.
    \item Quality and Size of Data (Generalizability and Interpretability): The dataset's size and quality, including missing data and outliers, are crucial in model selection. Larger datasets support more complex models, while smaller or poorer quality datasets might benefit from simpler models.
    \item Type of Features (Numerical and Categorical): The quantity and type of features (numerical and categorical) influence the model choice. For example, more categorical features might necessitate models adept at handling feature conversion.
    \item Nature of the Target Variable (Continuous): The target variable `selling\_price' is continuous, directing the choice towards regression models.
    \item Model Type (Linear vs. Non-linear): The decision between linear and non-linear models depends on the complexity of relationships within the data.
    \item Evaluation Criteria (Performance Metrics): Models are assessed based on RMSE, MAE, and R² score, with cross-validation for generalizability.
\end{enumerate}

Given these considerations, the preferred models are:
\begin{lstlisting}[frame=single, breaklines=true]
Random Forest Regressor: Effective for non-linear relationships and robust to unbalanced datasets. Requires hyperparameter tuning for optimal performance.
Gradient Boosting Regressors (e.g., XGBoost, LightGBM): High accuracy models suitable for complex data, but less interpretable and sensitive to overfitting.
Linear Regression: A baseline model offering high interpretability, suitable for linear relationships.
Ridge/Lasso Regression: Useful in the presence of multicollinearity, with regularization to counter overfitting.
Support Vector Regressor (SVR): Suitable for high-dimensional spaces, necessitating careful hyperparameter tuning.

The transition from one model to another is guided by evaluation metrics, with a preference for simpler models in cases of smaller or lower-quality datasets and a shift to more complex models as the data size and quality improve.
\end{lstlisting}


\subsection{Overall Representation: GPT Modeling Techniques}

\begin{figure*}[!ht]
\centering
\includegraphics[scale=0.78]{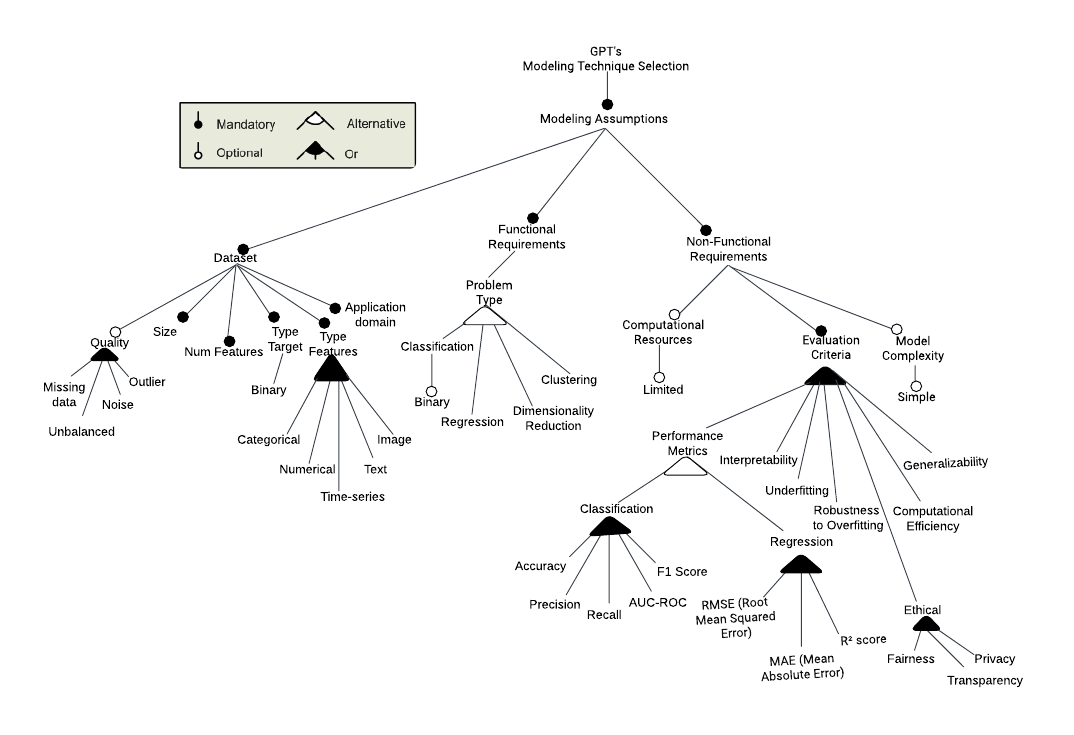}
\centering
\caption{Instance feature diagram of the proposed approach to represent GPT Modeling Assumptions feature diagram.}
\label{fig:approach}
\end{figure*}

Figure \ref{fig:approach} presents the feature diagram showcasing the factors and assumptions used in GPT model selection, integrating insights from both top-down and bottom-up approaches. It categorizes these into three main groups under Modeling Assumptions: dataset assumptions, functional requirements, and non-functional requirements, following the framework proposed by Tavares et al. \cite{tavares2022adaptive}. At the root of this diagram is the Modeling Technique Selection feature, which encompasses a wide range of machine learning techniques. Previous figures depicted only the ML techniques recommended for specific applications, whereas this comprehensive diagram includes all models for reference, as detailed in Tavares et al. \cite{tavares2022adaptive}.

Although ethical concerns were raised in the top-down query and we represent it in our Modeling Assumptions feature diagram, GPT did not consider these when proposing model selections for any of the three datasets.

Below is a concise set of constraints derived from the figure and GPT outputs, encapsulating dataset attributes, problem types, evaluation metrics, and computational capacity. The guidelines also consider the iterative process of model improvement, advocating for an adaptive approach—progressing from basic to advanced models in response to performance results and overfitting indicators. The following exemplifies the structuring of these constraints:

\begin{enumerate}
    \item Dataset Constraints:
    \begin{itemize}
        \item Size $\geq$ MinimumSizeRequirement $\land$ Features $\leq$ MaximumFeaturesAllowed
        \item DatasetType = \{Numerical, Categorical, Binary, Text, Image, Time-series\}
        \item DatasetQualityConstraints = $\neg$(MissingData $\vee$ Outliers $\vee$ Noise $\vee$ Unbalanced)
    \end{itemize}
    
\item Algorithm Type Constraints:
\begin{itemize}
    \item ProblemType = $\{$Classification $\vee$ Regression $\vee$ Clustering $\vee$ Dimensionality Reduction$\}$
    \item If Classification $\Rightarrow$ AlgorithmOptions = $\{$DecisionTree, NaiveBayes, NeuralNetwork, SVM, K-Nearest Neighbors, Logistic Regression, EnsembleMethods, DeepLearningModels$\}$
    \item If Regression $\Rightarrow$ AlgorithmOptions = $\{$LinearRegression, PolynomialRegression, RidgeRegression, LassoRegression, ElasticNet, EnsembleMethods, DeepLearningModels$\}$
    \item If Clustering $\Rightarrow$ AlgorithmOptions = $\{$K-Means, Hierarchical Clustering, DBSCAN, Gaussian Mixture Models, Mean Shift$\}$
    \item If Dimensionality Reduction $\Rightarrow$ AlgorithmOptions = $\{$PCA, t-SNE, LDA, QDA, Autoencoders$\}$
\end{itemize}

\item Evaluation Criteria Constraints:
\begin{itemize}
    \item If ProblemType = Classification $\Rightarrow$ PerformanceMetrics = \{Accuracy, Precision, Recall, F1Score, AUC-ROC\}
    \item If ProblemType = Regression $\Rightarrow$ PerformanceMetrics = \{RMSE, MAE, R$^2$\}
    \item If ProblemType = Clustering $\Rightarrow$ PerformanceMetrics = \{SilhouetteScore, DaviesBouldinIndex, CalinskiHarabaszIndex\}
    \item If ProblemType = DimensionalityReduction $\Rightarrow$ PerformanceMetrics = \{ReconstructionError, ExplainedVarianceRatio\}
\end{itemize}

\item Non-Functional Requirements Constraints:
\begin{itemize}
    \item EvaluationCriteria $\Rightarrow$ (Interpretability $\vee$ Robustness $\vee$ Fairness $\vee$ Privacy $\vee$ Transparency $\vee$ Generalizability $\vee$ EthicalConsiderations)
    \item ModelComplexity $\Rightarrow$ (Simple $\vee$ Complex)
\end{itemize}

    \item Computational Resource Constraints:
    \begin{itemize}
        \item If LimitedResources $\Rightarrow$ Prefer \{ModelsWithLessComplexity $\vee$ ReducedFeatureSets\}
        \item If SufficientResources $\Rightarrow$ Capability to Implement \{EnsembleMethods $\vee$ DeepLearningModels\}
    \end{itemize}
    
    \item Transition and Iterative Improvement Constraints:
    \begin{itemize}
        \item InitialModelSelection $\Rightarrow$ SimplerModels
        \item PoorPerformance $\Rightarrow$ TransitionToMoreComplexModels
        \item If OverfittingDetected $\Rightarrow$ ApplyRegularization $\vee$ ParameterTuning $\vee$ ModelSelectionReevaluation
    \end{itemize}
\end{enumerate}

In summary, this section presents the factors and variabilities that GPT takes into account when selecting a model for a dataset. This diagram and constraints can be further adapted to examine model selection for additional datasets, employing GPT's heuristics to interpret decisions made by an automated GPT-based system, or to conduct tests and verification processes investigating the relationships between these factors. 

\section{Comparative Results}

Tavares et al. \cite{tavares2022adaptive} present a system ranking models for datasets based on Scikit Heuristics. We tested this system on these three datasets to evaluate if the heuristic consistently selects similar models. When discrepancies occurred, we compared the top model from each heuristic.

We applied the main models suggested by the GPT and Scikit Learn diagrams using the Scikit Learn library. The GridSearchCV from scikit-learn was employed to automate the evaluation of all model parameter combinations via cross-validation.

\subsection{Heart Failure Prediction}
Following \cite{leenings2021photonai}, we divided our dataset into an 80\% training set (239 patients) and a 20\% test set (60 patients), using a stratified split to accommodate the dataset's imbalance.

The Sklearn heuristic recommended LinearSVC, KNeighborsClassifier, SVC, and EnsembleClassifiers. In contrast, the GPT heuristic suggested Logistic Regression, Random Forest, Gradient Boosting Machines, and Support Vector Machine. Model selection was based on performance metrics like accuracy, precision, recall, and AUC-ROC, as suggested by GPT.

We executed the following models for the dataset:
\begin{figure*}[ht!]
\centering
\caption{Executed Models for Dataset 01.}
\label{table:codev1}
\begin{lstlisting}[basicstyle=\scriptsize\ttfamily,numbers=none]
csv_path = 'dataset/gpt/heart_failure_clinical_records_dataset.csv'  
models_with_params = {
    LinearSVC: {'C': [0.01, 0.1, 1, 10]},
    KNeighborsClassifier: {'n_neighbors': [2, 3, 5, 7, 10]},
    SVC: {'C': [0.1, 1, 10], 'kernel': ['linear', 'rbf']},
    LogisticRegression: {'C': [0.1, 1, 2, 5, 10]},
    RandomForestClassifier: {'n_estimators': [2,5,10,100, 200], 'max_depth': [5, 10, None]},
    GradientBoostingClassifier: {'learning_rate': [0.01, 0.1, 0.2], 'n_estimators': [100, 200]}
}
metrics = ['accuracy', 'precision', 'recall', 'roc_auc']
evaluate_models_with_best_params(csv_path, models_with_params, metrics)
\end{lstlisting}
\end{figure*}

Results including the top two models from GPT (Logistic Regression and Random Forest), the top two from Scikit-learn (LinearSVC and KNeighborsClassifier), and the baseline model (Random Forest) as used in \cite{leenings2021photonai} are:

\begin{lstlisting}[frame=single, breaklines=true]
Best Results for LinearSVC with params {'C': 0.01}:
  accuracy: 0.8167
  precision: 0.7857
  recall: 0.5789
  roc_auc: 0.7529
  Best CV score: 0.8409
  Test score: 0.8167

Best Results for KNeighborsClassifier with params {'n_neighbors': 5}:
  accuracy: 0.7000
  precision: 0.5714
  recall: 0.2105
  roc_auc: 0.7561
  Best CV score: 0.7616
  Test score: 0.7000

Best Results for SVC with params {'C': 1, 'kernel': 'linear'}:
  accuracy: 0.7833
  precision: 0.8000
  recall: 0.4211
  roc_auc: 0.6861
  Best CV score: 0.8449
  Test score: 0.7833

Best Results for LogisticRegression with params {'C': 5}:
  accuracy: 0.8167
  precision: 0.7857
  recall: 0.5789
  roc_auc: 0.8588
  Best CV score: 0.8409
  Test score: 0.8167

Best Results for RandomForestClassifier with params {'max_depth': 10, 'n_estimators': 100}:
  accuracy: 0.8167
  precision: 0.7857
  recall: 0.5789
  roc_auc: 0.8896
  Best CV score: 0.8662
  Test score: 0.8167

Best Results for GradientBoostingClassifier with params {'learning_rate': 0.2, 'n_estimators': 100}:
  accuracy: 0.7833
  precision: 0.6875
  recall: 0.5789
  roc_auc: 0.8331
  Best CV score: 0.8620
  Test score: 0.7833
\end{lstlisting}

Logistic Regression, LinearSVC, and RandomForestClassifier showed identical accuracy and precision. However, Logistic Regression (suggested by GPT) demonstrated superior ROC-AUC performance. ROC-AUC is a critical metric for binary classifiers, especially in imbalanced datasets, as it quantifies the model's ability to distinguish between classes across all thresholds.

\subsection{Diabetes Prediction}
In the Diabetes Prediction subsection, we compared the effectiveness of models suggested by both Scikit-learn heuristics and GPT outputs. The Scikit-based heuristic recommended the SGDClassifier, and kernelApproximation, with alternatives like Linear SVC, Kneighbors Classifier, SVC, and Ensemble Classifiers for datasets under 100K samples. The GPT heuristic, on the other hand, suggested Logistic Regression, Decision Trees, Random Forests, GBM, and Neural Networks, emphasizing the transition between models based on performance metrics such as accuracy and AUC-ROC.

\begin{figure*}[ht!]
\centering
\caption{Executed Models for Dataset 02.}
\label{table:codev2}
\begin{lstlisting}[basicstyle=\scriptsize\ttfamily,numbers=none]
csv_path = 'dataset/gpt/diabetes_prediction_dataset.csv'  
models_with_params = {
    LogisticRegression: {'C': [0.1, 1, 2, 5, 10]},
    RandomForestClassifier: {'n_estimators': [2, 5, 10, 100, 200], 'max_depth': [5, 10, None]},
    GradientBoostingClassifier: {'learning_rate': [0.01, 0.1, 0.2], 'n_estimators': [100, 200]},
    SGDClassifier: {'alpha': [0.0001, 0.001, 0.01, 0.1], 'loss': ['hinge', 'log', 'modified_huber']},
    DecisionTreeClassifier: {'max_depth': [None, 5, 10, 15], 'min_samples_split': [2, 5, 10]},
}
metrics = ['accuracy', 'precision', 'recall', 'roc_auc']
evaluate_models_with_best_params(csv_path, models_with_params, metrics)
\end{lstlisting}
\end{figure*}

\begin{lstlisting}[frame=single, breaklines=true]
Best Results for LogisticRegression with params {'C': 0.1}:
  accuracy: 0.9604
  precision: 0.8592
  recall: 0.6388
  roc_auc: 0.9625
  Best CV score: 0.9602
  Test score: 0.9604

Best Results for RandomForestClassifier with params {'max_depth': 10, 'n_estimators': 10}:
  accuracy: 0.9721
  precision: 0.9931
  recall: 0.6771
  roc_auc: 0.9694
  Best CV score: 0.9718
  Test score: 0.9721

Best Results for GradientBoostingClassifier with params {'learning_rate': 0.1, 'n_estimators': 100}:
  accuracy: 0.9723
  precision: 0.9783
  recall: 0.6894
  roc_auc: 0.9794
  Best CV score: 0.9718
  Test score: 0.9723

  Best Results for SGDClassifier with params {'alpha': 0.001, 'loss': 'log'}:
  accuracy: 0.9599
  precision: 0.8598
  recall: 0.6312
  roc_auc: 0.9625
  Best CV score: 0.9605
  Test score: 0.9599

Best Results for DecisionTreeClassifier with params {'max_depth': 5, 'min_samples_split': 2}:
  accuracy: 0.9723
  precision: 1.0000
  recall: 0.6741
  roc_auc: 0.9541
  Best CV score: 0.9718
  Test score: 0.9723
\end{lstlisting}

Our execution and evaluation of these models on the dataset revealed notable results. The Logistic Regression model, recommended by GPT, achieved an accuracy of 0.9604, precision of 0.8592, recall of 0.6388, and a roc\_auc of 0.9625. In contrast, the SGDClassifier, as suggested by Scikit, showed similar accuracy (0.9599) and precision (0.8598), but slightly lower recall (0.6312) and identical roc\_auc (0.9625).

Notably, the RandomForestClassifier and GradientBoostingClassifier, also recommended by GPT, outperformed both in terms of accuracy, with scores of 0.9721 and 0.9723 respectively, and showed superior recall and precision. This comparison indicates that while the Scikit-based heuristic offers competitive models, the GPT-suggested models, particularly RandomForestClassifier and GradientBoostingClassifier, display a slight edge in overall performance for this specific dataset.

\subsection{Car's Price Prediction}
In the Car's Price Prediction subsection, we evaluated models suggested by both the Sklearn-based heuristic and GPT outputs. The Sklearn heuristic recommended RidgeRegression, SVR(kernel=linear), SVR(kernel=rbf), and EnsembleRegressors, whereas GPT suggested Random Forest Regressor, Gradient Boosting Regressors (like XGBoost, LightGBM), Linear Regression, Ridge/Lasso Regression, and Support Vector Regressor (SVR). Model performance was assessed based on RMSE, MAE, and R² score.

\begin{figure*}[ht!]
\centering
\caption{Executed Models for Dataset 03.}
\label{table:codev3}
\begin{lstlisting}[basicstyle=\scriptsize\ttfamily,numbers=none]
models_with_params = {
    Ridge: {'alpha': [0.1, 1, 10]},
    SVR: {'kernel': ['linear'], 'C': [0.1, 1, 10]},
    SVR: {'kernel': ['rbf'], 'C': [0.1, 1, 10], 'gamma': ['scale', 'auto']},
    RandomForestRegressor: {'n_estimators': [100, 200], 'max_depth': [5, 10, None]},
    GradientBoostingRegressor: {'learning_rate': [0.01, 0.1, 0.2], 'n_estimators': [100, 200]},
    LinearRegression: {},
    Lasso: {'alpha': [0.1, 1, 10]},
    SGDRegressor: {'alpha': [0.0001, 0.001, 0.01, 0.1]}
}
csv_path = 'dataset/gpt/CAR DETAILS FROM CAR DEKHO.csv'  
evaluate_regression_models_with_best_params(csv_path, models_with_params)
\end{lstlisting}
\end{figure*}

\begin{lstlisting}[frame=single, breaklines=true]
Best Results for Ridge with params {'alpha': 0.1}:
  RMSE: 339373.6684
  MAE: 122614.1828
  R2_score: 0.6226

Best Results for SVR with params {'C': 10, 'gamma': 'scale', 'kernel': 'rbf'}:
  RMSE: 568437.5465
  MAE: 288291.4370
  R2_score: -0.0588

Best Results for RandomForestRegressor with params {'max_depth': None, 'n_estimators': 200}:
  RMSE: 360838.8792
  MAE: 119192.8818
  R2_score: 0.5733

Best Results for GradientBoostingRegressor with params {'learning_rate': 0.2, 'n_estimators': 200}:
  RMSE: 354280.1772
  MAE: 146308.3123
  R2_score: 0.5887

Best Results for LinearRegression with params {}:
  RMSE: 340865.1113
  MAE: 120932.3265
  R2_score: 0.6193

Best Results for Lasso with params {'alpha': 10}:
  RMSE: 358057.2296
  MAE: 118037.6802
  R2_score: 0.5799

Best Results for SGDRegressor with params {'alpha': 0.0001}:
  RMSE: 355592.3303
  MAE: 148396.1888
  R2_score: 0.5857
\end{lstlisting}

Upon execution, the Ridge Regression model, recommended by Sklearn, achieved an RMSE of 339373.6684, MAE of 122614.1828, and an R2\_score of 0.6226. The SVR with RBF kernel, another Sklearn suggestion, however, had a significantly higher RMSE (568437.5465) and lower R2\_score (-0.0588).

Comparatively, the Random Forest Regressor, a GPT suggestion, showed an RMSE of 360838.8792 and an R2\_score of 0.5733, indicating slightly lower performance than Ridge Regression but notably better than SVR with RBF kernel. The GradientBoostingRegressor, also recommended by GPT, had an RMSE of 354280.1772 and an R2\_score of 0.5887, performing comparably to the Random Forest Regressor.

Linear Regression and Lasso, both GPT suggestions, showed competitive performance, with RMSEs of 340865.1113 and 358057.2296, respectively, and similar R2\_scores.

This comparison highlights that while Sklearn-based heuristics offer robust models like Ridge Regression, GPT's recommendations, especially GradientBoostingRegressor and Random Forest Regressor, are equally competitive in terms of RMSE and R2\_score. The overall performance of models suggested by GPT demonstrates their efficacy in handling this specific dataset for car price prediction.

\subsection{Discussion}

This section presents an overview of the results from evaluating the efficacy of model selection heuristics provided by GPT and Scikit-learn engineers across different datasets. Our findings reveal interesting implications for the application of Large Language Models (LLMs) like GPT-4 in data science. In the classification datasets, the models suggested by GPT outperformed those recommended by Scikit-learn heuristics.  Conversely, in the regression dataset, the model suggested by Scikit-learn engineers showed superior performance compared to GPT's suggestions. 

However, it's important to note that several factors can influence these outcomes. The way datasets are split, parameter selection for algorithms, and the specific characteristics of each dataset play significant roles in the performance of the suggested models. These factors need to be carefully considered when interpreting the results and their applicability to real-world scenarios.

Overall, the results are positive and encouraging. We were successful in mapping the decision-making process of GPT for the selection of each algorithm. The suggestions provided by GPT achieved satisfactory results in all datasets, demonstrating its potential as a valuable tool in model selection.

\section{Conclusion and Future Work}


This paper addressed the need for understanding the heuristics behind decision-making processes in AI, particularly in the context of model selection in data science. The increasing reliance on Large Language Models like GPT-4 in various data science applications underscores the importance of this endeavor. Our approach involved capturing and articulating the heuristics underlying GPT-4's model selection recommendations, a task crucial for enhancing transparency and trust in AI systems. It highlights the importance of various factors in the model selection process, including the nature of the data, problem type, performance metrics, computational resources, interpretability vs accuracy, assumptions about data, and ethical considerations.

To achieve this, we employed feature models, which proved instrumental in representing the complex, multi-faceted nature of these heuristics. By utilizing toy datasets, we were able to effectively test and evaluate both the model and the application of the captured heuristics. This methodology enabled us to compare the results against heuristics proposed by other platforms, providing a comprehensive assessment of the efficacy and uniqueness of GPT-4's approach.

\section{Future Work} \label{sec:threats}
 Our research opens several avenues for future exploration. One key area involves further refining the feature models to capture even more nuanced aspects of AI heuristics. Additionally, expanding the scope of datasets and scenarios will provide a more robust evaluation of AI decision-making across different contexts. There's also a compelling opportunity to explore the integration of these heuristics into real-world applications, assessing how they can enhance the performance and reliability of AI systems in practical settings. 
 
 Furthermore, the implications of this study for AI ethics and governance are vast and warrant deeper investigation. Understanding AI decision-making is a step towards more ethical and responsible AI, and future research should delve into how these insights can be translated into policy and practice. By continuing this line of inquiry, we aim to contribute to the development of AI systems that are not only powerful and efficient but also transparent, understandable, and aligned with human values and societal needs.

\subsection{Exploration of Other Large Language Models (LLMs)}
A comprehensive investigation of various LLMs will provide insights into their unique decision-making processes. Establishing clear evaluation standards will enable us to compare and contrast these models. The integration of different generative models with data connectors, as exemplified in LangChain, offers a fertile ground for exploring how different LLMs interact and perform in a variety of contexts.

\subsection{Prompt Engineering and Its Impact}
Prompt engineering and its influence on model heuristics is a significant aspect to explore. By adopting the approach proposed by \cite{aher2023using}, where LLMs are used to simulate multiple humans and replicate human subject studies, we can investigate how simulating data scientists with varied professional experiences impact GPT's model selection. This approach aligns with the concept of Turing Experiments for language models, which aim to evaluate the extent to which language models can simulate different aspects of human behavior.

\subsection{Ethical Considerations}
Our investigation revealed that while GPT's strategic framework (top-down approach) acknowledges ethical considerations, these elements are not as emphasized in the practical application of the models. This gap suggests an area for future work, where we can explore potential biases and ethical implications of AI-assisted model selection, and how prompt engineering could be leveraged to explicitly include these ethical considerations. By incorporating these factors, we can ensure that the AI’s decision-making process not only focuses on performance metrics but also aligns with ethical guidelines and best practices.

\subsection{Addressing Hallucination in Responses}
One of the critical challenges with LLMs is addressing `hallucinations' \cite{ji2023survey} – instances where the models generate contextually irrelevant or factually incorrect responses. A deeper analysis of the internal decision-making processes of these models is necessary to understand and mitigate these risks, which are crucial for the trust and safety in AI applications.

\subsection{Application to More Complex Scenarios}
Applying our findings to larger datasets and more complex applications will significantly enhance our understanding of LLM decision-making. This will involve not only scaling up the size of the datasets but also tackling more challenging scenarios that reflect real-world complexity. Such applications will provide valuable insights into the robustness and adaptability of LLMs in diverse and demanding environments.

\section*{Acknowledgment}
This work was supported by the Natural Sciences and Engineering Research Council of Canada (NSERC), and the Centre for Community Mapping (COMAP).
\bibliographystyle{IEEEtran}
\bibliography{references}

\end{document}